\documentclass{article}
\usepackage{ijcai21}
\usepackage{times}
\usepackage{soul}
\usepackage{epsfig}
\usepackage{url}
\usepackage[hidelinks]{hyperref}
\usepackage[utf8]{inputenc}
\usepackage[small]{caption}
\usepackage{graphicx}
\usepackage{amsmath}
\usepackage{amssymb}
\usepackage{amsthm}
\usepackage{booktabs}
\usepackage{algorithm}
\usepackage{algorithmic}
\usepackage{siunitx}
\usepackage{subcaption}
\newcommand\sbullet[1][.5]{\mathbin{\vcenter{\hbox{\scalebox{#1}{$\bullet$}}}}}
\newcommand\Tau{\mathcal{T}}
\title{Unsupervised Continual Learning Via Pseudo Labels}

\author{
Jiangpeng He \and
Fengqing Zhu \\
\affiliations
School of Electrical and Computer Engineering, Purdue University, West Lafayette, Indiana, USA\\
\emails
\{he416, zhu0\}@purdue.edu}

\begin{document}

\maketitle

\begin{abstract}
Continual learning aims to learn new tasks incrementally using less computation and memory resources instead of retraining the model from scratch whenever new task arrives. However, existing approaches are designed in supervised fashion assuming all data from new tasks have been manually annotated, which are not practical for many real-life applications. In this work, we propose to use pseudo label instead of the ground truth to make continual learning feasible in unsupervised mode. The pseudo labels of new data are obtained by applying global clustering algorithm and we propose to use the model updated from last incremental step as the feature extractor. Due to the scarcity of existing work, we introduce a new benchmark experimental protocol for unsupervised continual learning of image classification task under class-incremental setting where no class label is provided for each incremental learning step. Our method is evaluated on the CIFAR-100 and ImageNet (ILSVRC) datasets by incorporating the pseudo label with various existing supervised approaches and show promising results in unsupervised scenario. 
\end{abstract}

%%%%%%%%% BODY TEXT

\section{Introduction}
\label{introduction}
\begin{figure}[t]
\begin{center}
  \includegraphics[width=1.\linewidth]{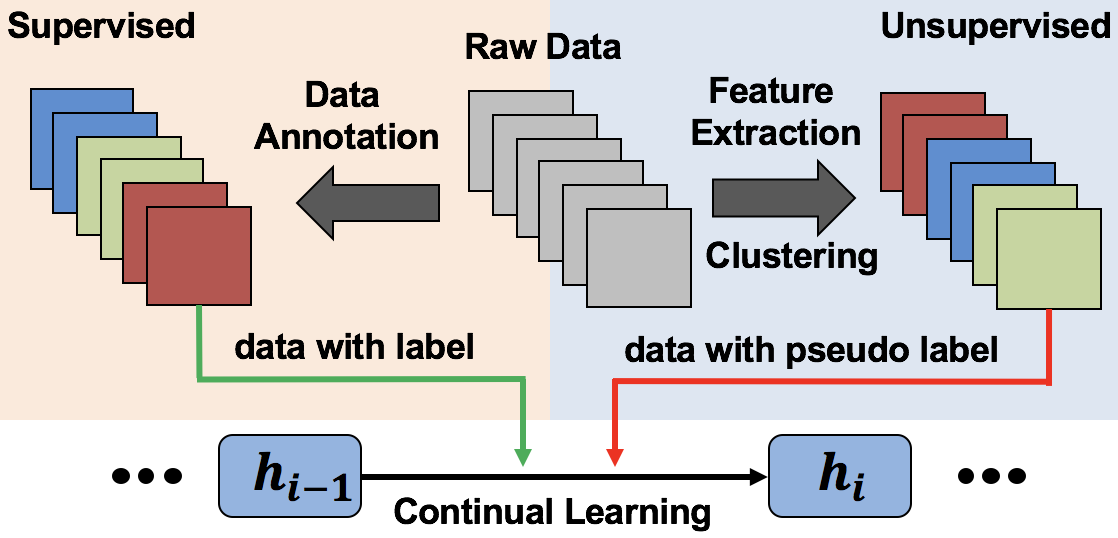}
  \vspace{-0.7cm}
  \caption{\textbf{Supervised vs. unsupervised continual learning for the new task i}. $\textbf{h}$ refers to the model in different incremental steps. The supervised and our proposed pseudo label based unsupervised continual learning are illustrated by green and red arrows respectively.}
  \label{fig:intro_diff}
\end{center}
\end{figure}

The success of many deep learning techniques rely on the following two assumptions: 1) training data is identically and independently distributed (\textit{i.i.d.}), which rarely happens if new data and tasks arrive sequentially over time, 2) labels for the training data are available, which requires additional data annotation by human effort, and can be noisy as well. Continual learning has been proposed to tackle issue \#1, which aims at learning new tasks incrementally without forgetting the knowledge on all tasks seen so far. Unsupervised learning focuses on addressing issue \#2 to learn visual representations used for downstream tasks directly from unlabeled data. However, unsupervised continual learning, which is expected to tackle both issues mentioned above, has not been well studied~\cite{survey_2020}. Therefore, we introduce a simple yet effective method in this work that can be adapted by existing supervised continual learning approaches in unsupervised setting where no class label is required during the learning phase. We focuses on image classification task under the class-incremental setting~\cite{hsu2018re} and the objective is to learn from unlabeled data for each incremental step while providing semantic meaningful clusters on all classes seen so far during inference. Figure~\ref{fig:intro_diff} illustrates the difference between the typical supervised and proposed unsupervised continual learning scenarios to learn a new task $i$. 
% that there is no repeated class in new tasks.
% Figure illustrates the difference between our method and exiting methods for class-incremental learning. 

Current continual learning approaches can be generally summarized into three categories including (1) \textit{Regularization based}, (2) \textit{Bias-correction based} and (3) \textit{Rehearsal based}. Our proposed method can be directly embedded into existing supervised approaches in category (1) and (2) with an additional step to extract features of unlabeled data and perform clustering to obtain pseudo label. However, for methods in (3), selecting exemplars from learned tasks when class label is not provided in unsupervised scenario is still an unsolved and challenging step. In this work, we tackle this issue by sampling the unlabeled data from the centroid of each generated cluster as exemplars to incorporate with \textit{Rehearsal based} approaches. 

Pseudo label~\cite{lee2013pseudolabel} is widely applied in both semi-supervised and unsupervised learning scenarios to handle unlabeled data for downstream tasks, which is effective due to its simplicity, generality and ease of implementation. However, whether it is feasible for continual learning to rely on pseudo labels instead of human annotations is not well unexplored yet, which is more challenging as we also need to address catastrophic forgetting~\cite{CF} in addition to learning new knowledge from unlabeled data.

In this work, we adopt K-means~\cite{K-MEANS} as our global clustering algorithm for illustration purpose and we propose to use the continual learning model (except the last fully connected layers) at every incremental step for feature extraction of unlabeled data to obtain pseudo label. The exemplars used for \textit{Rehearsal based} approaches are selected after applying k-means from each generated cluster based on the distance to cluster centroid without requiring the class labels. Note that we are not proposing new approach to address catastrophic forgetting for continual learning in this work, but instead we test the effectiveness of using pseudo labels to make existing supervised methods feasible in unsupervised setting. Therefore, we incorporate our method with existing representative supervised approaches from all three categories mentioned above including LWF~\cite{LWF}, ICARL~\cite{ICARL}, EEIL~\cite{EEIL}, LUCIR~\cite{rebalancing}, WA~\cite{mainatining} and ILIO~\cite{ILIO}. We show promising performance in unsupervised scenario on both CIFAR-100~\cite{CIFAR} and ImageNet (ILSVRC)~\cite{IMAGENET1000} datasets compared with results in supervised case that do require the ground truth for continual learning. The main contributions of this paper are summarized as follows.
\begin{itemize}

  \item We explore a novel problem for continual learning using pseudo labels instead of human annotations, which is both challenging and meaningful for real-life applications.

  \item Our proposed method can be easily adapted by existing supervised continual learning techniques and we achieve competitive performance on both CIFAR-100 and ImageNet in unsupervised scenario.
  
  \item A new benchmark evaluation protocol is introduced for future research work and extensive experiments are conducted to analyze the effectiveness of each component in our proposed method.
    
\end{itemize}

\section{Related Work}
\label{related work}
\subsection{Continual Learning}
The major challenge for continual learning is catastrophic forgetting~\cite{CF} where the model quickly forgets already learned knowledge due to the unavailability of old data during the learning phase of new tasks. Many effective techniques have been proposed to address catastrophic forgetting in supervised scenario, which can be divided into three main categories: (1) \textit{Regularization based} methods aim to retain old knowledge by constraining the change of parameters that are important for old tasks. Knowledge distillation loss~\cite{KD} is one of the representatives, which was first applied in~\cite{LWF} to transfer knowledge using soft target distribution from teacher model to student model. Later the variants of distillation loss proposed in~\cite{rebalancing,ILIO} are shown to be more effective by using stronger constraints. (2) \textit{Bias-correction based} strategy aims to maintain the model performance by correcting the biased parameters towards new tasks in the classifier. Wu \textit{et al}.~\cite{BIC} proposed to apply an additional linear layer with a validation sets after each incremental step. Weight Aligning (WA) is proposed in~\cite{mainatining} to directly correct the biased weights in the FC layer, which does not require extra parameters compared with previous one. (3) \textit{Rehearsal based} methods~\cite{ICARL,EEIL} use partial data from old tasks to periodically remind model of already learned knowledge to mitigate forgetting. 

However, all these methods require class label for the continual learning process, which limits their applications in real world. Therefore, in this work we propose to use pseudo label obtained from cluster assignments to make existing supervised approaches feasible in unsupervised mode.

\subsection{Unsupervised Representation Learning}
Many approaches have been proposed to learn visual representation using deep models with no supervision. \textit{Clustering} is one type of unsupervised learning methods that has been extensively studied in computer vision problems~\cite{deep_clustering,zhan2020online_cluster4}, which requires little domain knowledge from unlabeled data compared with self-supervised learning~\cite{jing2020self_survey}. Caron \textit{et al.}~\cite{deep_clustering} proposed to iteratively cluster features and update model with subsequently assigned pseudo labels obtained by applying standard clustering algorithm such as K-means~\cite{K-MEANS}. The most recent work~\cite{zhan2020online_cluster4} propose to perform clustering and model update simultaneously to address the model's instability during training phase. However, all these existing methods only work on static datasets and are not capable of learning new knowledge incrementally. In addition, the idea of using pseudo label is also rarely explored under continual learning context where the learning environment changes a lot since we need to address catastrophic forgetting as well besides learning visual representation from unlabeled data. In this work, we propose to use the fixed pseudo label for unsupervised continual learning which is described in Section~\ref{method_our_overview}. We also show in Section~\ref{exp:ablation} that iteratively perform clustering to update pseudo labels will result in performance degradation under continual learning context.

\section{Problem Setup}
\label{Sec: Continual Learning From Unlabeled Data}
Continual learning aims to learn knowledge from a sequence of new tasks. Broadly speaking, it can be divided into (1) task-incremental, (2) domain-incremental, and (3) class-incremental as discussed in~\cite{hsu2018re}. Methods designed for task-incremental problems use a multi-head classifier for each independent task and domain-incremental methods aim to learn the label shift instead of new classes. In this work, we study the unsupervised scenario under class-incremental setting, which is also known as Single-Incremental-Task~\cite{SIT} using a single-head classifier for inference. Specifically, the class-incremental learning problem $\Tau$ can be formulated as learning a sequence of $N$ tasks $\{\Tau^1,...,\Tau^N\}$ corresponding to $N-1$ incremental steps since the learning of the first task $\Tau^1$ is not related to incremental regime as no previous knowledge need to maintain. Each task $\Tau^i \in \Tau$ for $i \in \{1,...N\}$ contains $M^i$ non-overlapped new classes to learn. In this work, we study class-incremental learning in unsupervised scenario starting from task $\Tau^2$ for incremental steps only where we assume the initial model is supervisedly trained on $\Tau^1$. 
% Note that the first learning step $\Tau^1$ is not related to class-incremental learning as there is no learned knowledge we need to maintain.  
% we assume that there is an initial model $\textbf{h}_0$ already trained on $\Tau^0$ before incremental learning, which is the initial knowledge we need to pre.
Let $\{D^1, ..., D^N\}$ denotes the training data corresponds to $N$ tasks, where $D^i$ indicates the data belonging to the task $i$. In supervised case, $D^i = \{(\textbf{x}_1^i,y_1^i)...(\textbf{x}_{n_i}^i,y_{n_i}^i)\}$ where $\textbf{x}$ and $y$ represent the data and the label respectively, and $n_i$ refers to the number of total training data in $D^i$. In unsupervised case, we assume the labels of data are unknown for each incremental step so $D^i = \{\textbf{x}_1^i...\textbf{x}_{n_i}^i\}$ for $i\in\{2,3,...N\}$. The objective is to learn from unlabeled data for each incremental step while providing semantic meaningful clusters after each step on test data belonging to all classes seen so far. 

\textbf{Fixed step size: }As shown above $M^i$ refers to the number of added new classes for task $\Tau^i \in \Tau$, which is also defined as incremental step size . Existing benchmark protocols~\cite{ICARL,ILIO,rebalancing} for supervised continual learning use a fixed step size $M$ over all tasks where $M^i = M$ for $i \in {1,...N}$ and the continual learning under variable step size is not well studied yet even in supervised case. Therefore, we also assume that the number of new added classes for each task remain unchanged over the entire continual learning process, \textit{i.e.} the fixed step size $M$ is known in advance in our unsupervised setting while class labels of data in each incremental step are not provided. 

\textbf{Online and offline implementation: }Based on training restriction, continual learning methods can be implemented as either online or offline where the former methods~\cite{ILIO} use each data only once to update the model and the data can be used for multiple times in the offline case. In general, the online scenario is more closer to real life setting but is also more challenging to realize. In this work, our proposed method is implemented in both online and offline scenarios for unsupervised continual learning. Note that for our implementation in online case, we assume that we have access to all training data $\{\textbf{x}_1^i...\textbf{x}_{n_i}^i\} \in D^i$ before the learning of each new task $i$ but we use each data only once to update the model.

\begin{algorithm}[tb]
\caption{Unsupervised Continual Learning}
\label{alg:unsupervised continual learning}
\hspace*{0.02in} {\bf Input:}
a sequence of $N$ tasks $\{\Tau^1,...,\Tau^N\}$\\
\hspace*{0.02in} {\bf Input:} 
An initial model $\textbf{h}_0$ \\
\hspace*{0.02in} {\bf Require:} 
Clustering algorithm \textbf{$\Theta$}\\
\hspace*{0.02in} {\bf Output:} 
Updated model $\textbf{h}_N$
\begin{algorithmic}[1]
\STATE $M^1 \leftarrow |\Tau^1|_{class}$ \COMMENT{Added classes in first task}
\STATE
$\textbf{h}_1 \leftarrow \textit{Learning}(\Tau^1, \textbf{h}_0)$ \COMMENT{Learning first task}
\FOR{i = 2, ..., N}
\STATE $M^i \leftarrow |\Tau^i|_{class} $\COMMENT{number of new added classes} 
\STATE $D^i \leftarrow \{\textbf{x}_1, ..., \textbf{x}_{n_i}\}$ \COMMENT{Unlabeled training data in $\Tau^i$} 
\STATE $\textbf{h}_{fe} \leftarrow \textbf{h}_{i-1}$ \COMMENT{Feature extractor} 
\STATE \{$\tilde{a}_1$,.. $\Tilde{a}_{n_i}$\} $\leftarrow$ $\Theta (\textbf{h}_{fe}(D^i))$ \COMMENT{Cluster assignments} 
\STATE $\tilde{Y}^i \leftarrow \{\underset{_{k = 1,...n_i}}{\tilde{y}_k} = \tilde{a}_k + \sum\limits_{j=1}^{i-1} M^j$\} \COMMENT{Pseudo label}
\STATE $\textbf{h}_i \leftarrow \textit{Continual Learning}(D^i, \tilde{Y}^i, \textbf{h}_{i-1})$ 
\ENDFOR
\STATE \textbf{return} $h^N$ 

\end{algorithmic}
\end{algorithm}

% \input{method_existing}

%\vspace{-0.4cm}
\section{Our Method}
\label{method:our}
\label{method_our_overview}
In this work, we propose a simple yet effective method for unsupervised continual learning using pseudo label obtained based on cluster assignments. The overall procedure to learn a sequence of new tasks $\{\Tau^1,...,\Tau^N\}$ are illustrated in Algorithm~\ref{alg:unsupervised continual learning}. The updated model after learning each task is evaluated to provide semantic meaningful clusters on all classes seen so far. 

For illustration purpose, we adopt k-means as our global clustering algorithm to generate cluster assignments and obtain pseudo label, which will be illustrated in Section~\ref{method:obtain pseudo label}. Then, we demonstrate how to easily incorporate our method with existing supervised approaches in Section~\ref{method:continual learning}.

% knowledge distillation loss and exemplar set are adopted as our baseline solution. We also examine recently proposed techniques designed for supervised continual learning methods and show they can be incorporated into our framework for additional performance improvement. As shown in Figure~\ref{fig:method}, for each incremental step, we first extract feature embeddings for clustering using the current model except the last fully connected layer.  We then obtain pseudo label based on cluster assignments for continual learning by training with exemplars and knowledge distillation loss. After each incremental step, the updated model is used for feature extraction on the following task.

\subsection{Clustering: Obtain Pseudo Label}
\label{method:obtain pseudo label}
Clustering is one of the most common methods for unsupervised learning, which requires little domain knowledge compared with self-supervised techniques. We focus on using a general clustering method such as K-means~\cite{K-MEANS}, while we also provide the experimental results using other clustering methods as illustrated in \textit{Appendix}, which indicates that the choice is not critical for continual learning performance in our setting. Specifically, K-means algorithm learns a centroid matrix $\textbf{\textit{C}}$ together with cluster assignments $\tilde{a}_k$ for each input data $\textbf{x}_k$ by iteratively minimizing $\frac{1}{N}\sum_{k=1}^{N} ||\textbf{h}_{fe}(\textbf{x}_k) - \textbf{\textit{C}}\tilde{a}_k||_2^2$,
%Equation~\ref{eq:kmeans}
% \begin{equation}
%     \frac{1}{N}\sum_{k=1}^{N} ||\textbf{h}_{fe}(\textbf{x}_k) - \textbf{\textit{C}}\tilde{a}_k||_2^2
%     \label{eq:kmeans}
% \end{equation}
where $\textbf{h}_{fe}$ refers to the feature extractor. Let $m$ and $n$ represent the number of learned classes and new added classes respectively, then we have $\tilde{a}_k \in \{1, 2, ..., n\}$ and the pseudo label $\tilde{Y}$ for continual learning is obtained by $\{\tilde{y}_k  = \tilde{a}_k + m | k= 1,2,.. \}$ and $\tilde{y}_k \in \{m+1, m+2, ..., m+n\}$. 

Learning visual representation from unlabeled data using pseudo label is proposed in~\cite{deep_clustering}, which iteratively performs clustering and updating the feature extractor. However, they are not capable of learning new classes incrementally and the learning environment changes under continual learning context as we need to maintain the learned knowledge as well as learning from new tasks. Therefore, in this work we propose to apply the model, $\textbf{h}_{fe} = \textbf{h}_{i-1}$, obtained after incremental step $i-1$ (except the last fully connected layer) as the feature extractor for incremental step $i$ to extract feature embeddings on all unlabeled data belonging to the new task. Next, we apply k-means based on extracted features to generate cluster assignments and use the fixed pseudo label $\tilde{Y}$ to learn from new task during the entire incremental learning step $i$. We show in our experiments later that alternatively performing clustering and use pseudo label to update the model as in~\cite{deep_clustering} will result in performance degradation which is discussed in Section~\ref{exp:ablation}. Note that we assume $\textbf{h}_1$ is obtained from $\Tau^1$ in supervised mode as illustrated in Section~\ref{Sec: Continual Learning From Unlabeled Data}, so in this work we mainly focus on how to incrementally learn new classes from unlabeled data while maintaining performance on all old classes seen so far. 

\subsection{Incorporating into Supervised Approaches}
\label{method:continual learning}

The obtained pseudo label $\tilde{Y}$ can be easily incorporated with \textit{Regularization-based} methods using knowledge distillation loss or its variants. The distillation loss is formulated by Equation~\ref{eq:kdloss}
\begin{equation} \label{eq:kdloss}
\begin{aligned} 
L_{D} = \frac{1}{N}\sum_{k=1}^N\sum_{r=1}^{m}-\hat{p}_T^{(r)}(\textbf{x}_k)log[p_T^{(r)}(\textbf{x}_k)]
\end{aligned}
\end{equation}
% \vspace{-0.4cm}
$$
\begin{aligned}
\hat{p}_T^{(r)} =
\frac{\exp{(\hat{o}^{(r)}}/T)}{\sum_{j=1}^m\exp{(\hat{o}^{(j)}}/T)}\ , \ 
p_T^{(r)} = \frac{\exp{(o^{(r)}}/T)}{\sum_{j=1}^m\exp{(o^{(j)}}/T)}
\end{aligned}
$$
where $\hat{o}^{m\times 1}$ and $o^{m\times 1}$ denote the output logits of student and teacher models respectively for the $m$ learned classes. $T$ is the temperature scalar used to soften the probability distribution. The cross entropy loss to learn the added $n$ new classes can be expressed as
\begin{equation}
    L_{C} = \frac{1}{N}\sum_{k=1}^N\sum_{r=1}^{n+m}-\tilde{y}_k^{(r)}log[p^{(r)}(\textbf{x}_k)]
\end{equation}
where $\tilde{y}_k \in \tilde{Y}$ is the obtained pseudo label for data $\textbf{x}_k$ instead of the ground truth labels in supervised case. Then the cross-distillation loss combining cross entropy $L_{C}$ and distillation $L_{D}$ is formulated in Equation~\ref{eq:cdloss} with a hyper-parameter $\alpha = \frac{m}{m+n}$ to tune the influence between two terms.  
\begin{equation} \label{eq:cdloss}
\begin{aligned}
L_{CD}(\textbf{x}) = \alpha L_D(\textbf{x}) + (1-\alpha) L_C(\textbf{x})
\end{aligned}
\end{equation} 

Herding dynamic algorithm~\cite{HERDING} is widely applied for \textit{Rehearsal based} methods to select exemplars based on class mean in supervised case. However, since no class label is provided in unsupervised scenario, we instead propose to select exemplars based on cluster mean. Algorithm~\ref{alg:exp_selection} describes exemplar selection step for task $\Tau^i$. The exemplar set $Q$ stores the data and pseudo label pair denoted as $(\textbf{x}_k, \tilde{y}_k)$. 

The incorporation with \textit{Bias-correction based} methods is the most straightforward. BIC~\cite{BIC} applies an additional linear model for bias correction after each incremental step using a small validation set containing balanced old and new class data. In our unsupervised scenario, both the training and validation set used to estimate bias can be constructed using obtained pseudo label instead of the ground truth. The most recent work WA~\cite{mainatining} calculates the norms of weights vectors in FC layer for old and new class respectively and use the ratio to correct bias without requiring extra parameters. Thus our method can be directly embedded with it by an addition step to obtain pseudo label as illustrated in Section~\ref{method:obtain pseudo label}. 

We emphasize that we are not introducing new method to address catastrophic forgetting, but rather investigating whether it is possible to use pseudo labels instead of ground truth labels for continual learning. We show in Section~\ref{experimental results} that our proposed method works effectively with existing approaches from all categories mentioned above.

\begin{algorithm}[tb]
\caption{Unsupervised Exemplar Selection }
\label{alg:exp_selection}
\hspace*{0.02in} {\bf Input:}
image set $D^i = \{\textbf{x}_1,...,\textbf{x}_{ni}\}$ from task $\Tau^i$\\
\hspace*{0.02in} {\bf Input:} 
\textit{q} target exemplars per class \\
\hspace*{0.02in} {\bf Require:} 
clustering algorithm \textbf{$\Theta$}\\
\hspace*{0.02in} {\bf Require:} 
feature extractor $\textbf{h}_{fe} = \textbf{h}_i$\\
\hspace*{0.02in} {\bf Output:} 
Exemplar set \textit{Q}
\begin{algorithmic}[1]
\STATE $M^i \leftarrow |\Tau^i|_{class} $\COMMENT{number of new added classes}
\STATE \{$\tilde{a}_1$,.. $\Tilde{a}_{n_i}$\} $\leftarrow$ $\Theta (\textbf{h}_{fe}(D^i))$ \COMMENT{Cluster assignments}
\FOR{j = 1, 2,..., $M_i$}
\STATE $\textbf{X}_j \leftarrow \{\textbf{x}_n| \tilde{a}_n = j\}$
\STATE $\mu_j \leftarrow \frac{1}{|\textbf{X}_j|}\sum_{\textbf{x}\in \textbf{X}_j} \textbf{h}_{fe}(\textbf{x})$ \COMMENT{Cluster mean}
\FOR{k = 1, 2,..., \textit{q}}
\STATE $\textbf{e}_k \leftarrow \underset{\textbf{x}\in \textbf{X}_j}{\textit{argmin}}\{\mu_j - \frac{1}{k}[\textbf{h}_{fe}(\textbf{x}) + \sum_{l=1}^{k-1}\textbf{h}_{fe}(\textbf{e}_l)]\}$ \COMMENT{herding selection~\cite{HERDING} within cluster}
\ENDFOR
\STATE $Q \leftarrow Q \cup \{\textbf{e}_1, ..., \textbf{e}_q\}$
\ENDFOR
\STATE \textbf{return} $Q$ 
\end{algorithmic}
\end{algorithm}

\begin{table*}[t]
    % \vspace{-0.3cm}
    \centering
    \scalebox{.9}{
    \begin{tabular}{crrrrrrrr|rrrr}
        \hline
        Datasets & \multicolumn{8}{c}{\textbf{CIFAR-100}} & \multicolumn{4}{|c}{\textbf{ImageNet}} \\
        \hline
        Step size& \multicolumn{2}{c}{5} & \multicolumn{2}{c}{10} & \multicolumn{2}{c}{20} & \multicolumn{2}{c}{50} & \multicolumn{2}{|c}{10} & \multicolumn{2}{c}{100}\\
        \hline
        ACC & \multicolumn{1}{c}{Avg} & \multicolumn{1}{c}{Last} & \multicolumn{1}{c}{Avg} & \multicolumn{1}{c}{Last} & \multicolumn{1}{c}{Avg} & \multicolumn{1}{c}{Last} & \multicolumn{1}{c}{Avg} &
        \multicolumn{1}{c}{Last} & \multicolumn{1}{|c}{Avg} &
        \multicolumn{1}{c}{Last} & \multicolumn{1}{c}{Avg} &
        \multicolumn{1}{c}{Last} \\
        \hline
        \hline
        LWF (w/) & 0.299 & 0.155& 0.393& 0.240& 0.465 & 0.352 & 0.512 & 0.512 & 0.602 & 0.391 & 0.528 & 0.374\\
        LWF+Ours (w/o, $\Delta$)  & -0.071 & -0.029 & -0.091& -0.025& -0.086 & -0.062 & -0.095 & -0.095 & \textbf{-0.033} & -0.053 & -0.211 & -0.174\\
        \hline
         ICARL (w/) & 0.606 & 0.461& 0.626& 0.518& 0.641 & 0.565 & 0.607 & 0.607 & 0.821 & 0.644 & 0.608 & 0.440\\
        ICARL+Ours (w/o, $\Delta$)  & -0.084 & -0.045& -0.135& -0.142& -0.158 & -0.174 & -0.108 & -0.108 & \textbf{-0.043} & -0.047 & -0.197 & -0.015\\
        \hline
        EEIL (w/) & 0.643 & 0.482& 0.638& 0.517& 0.637 & 0.565 & 0.603 & 0.603 & 0.893 & 0.805 & 0.696 & 0.520\\
        EEIL+Ours (w/o, $\Delta$)  & -0.071 & -0.043& -0.131& -0.121& -0.131 & -0.148 & -0.088 & -0.088 & \textbf{-0.040} & -0.064 & -0.199 & -0.154\\
        \hline
        LUCIR (w/) & 0.623 & 0.478& 0.631& 0.521& 0.647 & 0.589 & 0.642 & 0.642 & 0.898 & 0.835 & 0.834 & 0.751\\
        LUCIR+Ours (w/o, $\Delta$)  & \textbf{-0.015} & -0.003 & -0.104& -0.106& -0.131 & -0.152 & -0.111 & -0.111 & \textbf{-0.037} & -0.083 & -0.293 & -0.342\\
        \hline
        WA (w/) & 0.643 & 0.496& 0.649& 0.535& 0.669 & 0.592 & 0.655 & 0.655 & 0.905 & 0.841 & 0.859 & 0.811\\
        WA+Ours (w/o, $\Delta$)  & \textbf{-0.034} & -0.014& -0.110& -0.106& -0.121 & -0.136 & -0.092 & -0.092 & \textbf{-0.037} & -0.056 & -0.295 & -0.376\\
        \hline
        ILIO (w/) & 0.664 & 0.515& 0.676& 0.564& 0.681 & 0.621 & 0.652 & 0.652 & 0.903 & 0.845 & 0.696 & 0.601\\
        ILIO+Ours (w/o, $\Delta$)  & -0.123 & -0.194 & -0.140 & -0.175& -0.134& -0.157 & -0.106 & -0.106 & -0.057 & -0.118 & -0.178 & -0.212\\
        
        \hline
    \end{tabular}
    }
    \vspace{-0.2cm}
    \caption{\textbf{Summary of unsupervised results and the comparison with supervised case}. The average ACC (Avg) over all incremental steps and the last step ACC (Last) are reported. w/ and w/o denote with or without label for continual learning, \textit{i.e. }supervised or unsupervised. $\{\Delta = w/ - w/o\}$ shows the performance difference. Spotlight results ($|\Delta| < 0.05$) for Avg accuracy are marked in bold.}
    \label{tab:incorpor}
\end{table*}
\begin{figure*}[t!]
\vspace{-0.4cm}
\centering
\begin{minipage}[t]{0.22\linewidth}
    \centering
    \includegraphics[width=4.cm,height=3.5cm]{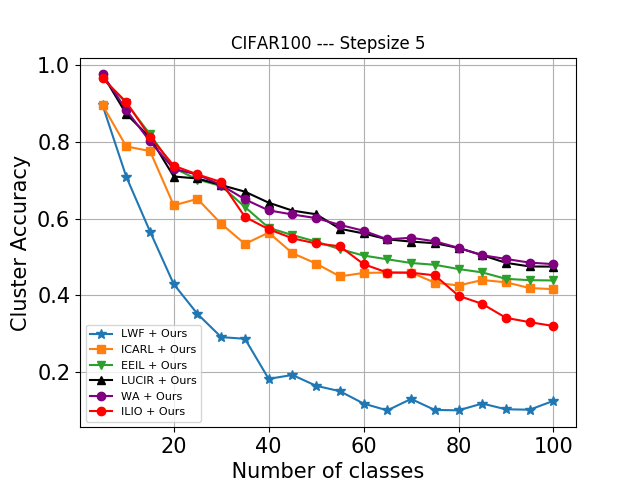}
    \parbox{15.5cm}{\small \hspace{2.cm}(a)}
\end{minipage}
\hspace{2ex}
\begin{minipage}[t]{0.22\linewidth}
    \centering
    \includegraphics[width=4.cm,height=3.5cm]{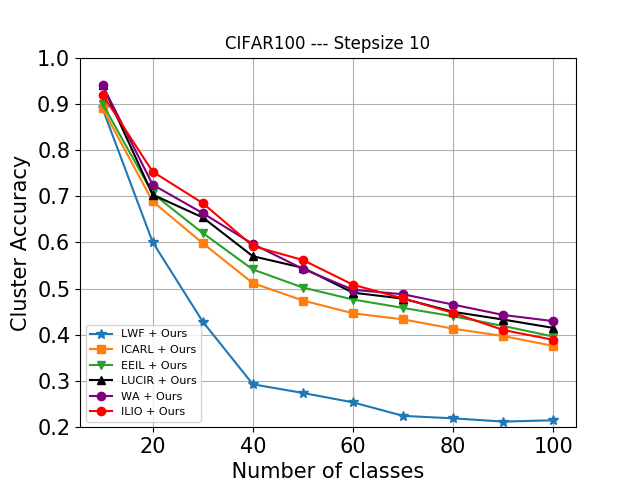}
    \parbox{15.5cm}{\small \hspace{2.cm}(b)}
\end{minipage}
\hspace{2ex}
\begin{minipage}[t]{0.22\linewidth}
    \centering
    \includegraphics[width=4.cm,height=3.5cm]{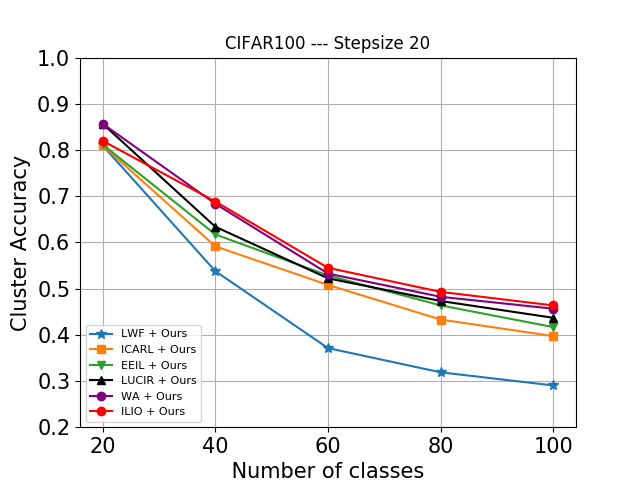}
    \parbox{15.5cm}{\small \hspace{2.cm}(c)}
\end{minipage}
\hspace{2ex}
\begin{minipage}[t]{0.22\linewidth}
    \centering
    \includegraphics[width=4.cm,height=3.5cm]{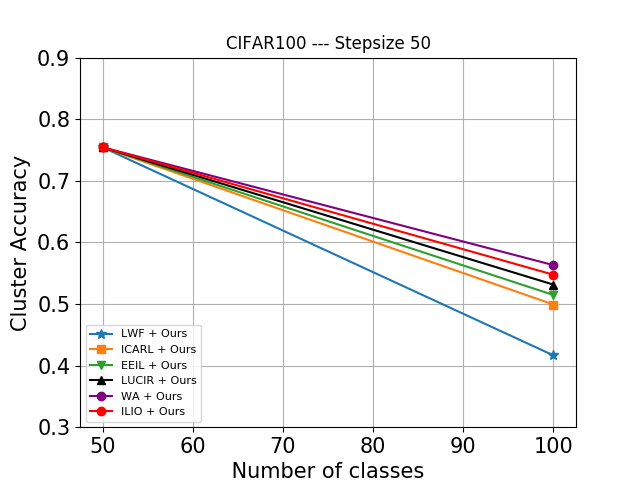}
    \parbox{15.5cm}{\small \hspace{2.cm}(d)}
\end{minipage}
\begin{center}
\vspace{-.45cm}
\caption{\textbf{Results on CIFAR-100} with step size 5, 10, 20, and 50 by incorporating our method with existing approaches to realize continual learning in unsupervised scenario. (Best viewed in color)}
\label{fig:result on cifar}
\end{center}
% \vspace{-.3cm}
\end{figure*}

% \begin{figure}[t]
% \begin{center}
% %   \vspace{-0.5cm}
%   \includegraphics[width=1.\linewidth]{fig_experiment/imagenet_new.png}
%   \vspace{-0.6cm}
%   \caption{\textbf{Results on ImageNet} with step size (a) 10 on ImageNet-100 and (b) 100 on ImageNet-1000.(Best viewed in color)}
% %   \vspace{-.3cm}
%   \label{fig:imagenet}
% \end{center}
% \end{figure}

\vspace{-0.15cm}
\section{Experimental Results}
\vspace{-0.15cm}
\label{experimental results}
In this section, we evaluate our proposed method from two perspectives. 1) We incorporate with existing approaches and compare results obtained in unsupervised and supervised cases to show the ability of using pseudo labels for unsupervised continual learning to provide semantic meaningful clusters for all classes seen so far. 2) We analyze the effectiveness of each component in our proposed method including the exemplar selection and the choice of feature extractor in unsupervised scenario. These experimental results are presented and discussed in Sections~\ref{results: compare with supervised} and \ref{exp:ablation}, respectively. (Additional results are available in \textit{Appendix}) 

\subsection{Benchmark Experimental Protocol}
\label{sec:protocol}
Although different benchmark experimental protocols are used in supervised case~\cite{ICARL,rebalancing,ILIO}, there is no agreed protocol for evaluation of unsupervised continual learning methods. In addition, various learning environments may happen when class label is not available so it is impossible to use one protocol to evaluate upon all potential scenarios. Thus, our proposed new protocol focuses on class-incremental learning setting and aims to evaluate the ability of unsupervised methods to learn from unlabeled data while maintaining the learned knowledge during continual learning. Specifically, the following assumptions are made: (1) all the new data belong to new class, (2) the number of new added class (step size) is fixed and known beforehand, (3) no class label is provided for learning (except for the initial step) and (4) the updated model should be able to provide semantic meaningful clusters for all classes seen so far during inference. Our protocol is introduced based on current research progress for supervised class-incremental learning and three benchmark datasets are considered including (i) CIFAR-100~\cite{CIFAR} with step size 5, 10, 20, 50 (ii) ImageNet-1000 (ILSVRC)~\cite{IMAGENET1000} with step size 100 and (iii) ImageNet-100 (100 classes subset of ImageNet-1000) with step size 10. Top-1 and Top-5 ACC are used for CIFAR-100 and ImageNet, respectively. 

\subsection{Evaluation Metrics}
We evaluate our method using cluster accuracy (ACC), which is widely applied in unsupervised setting~\cite{deep_clustering,SCAN:van2020scan} when class label is not provided. We first find the most represented class label for each cluster using Hungarian matching algorithm~\cite{kuhn1955hungarian}, and then calculate the accuracy as $\frac{N_{c}}{N}$ where $N$ is the total number of data and $N_{c}$ is the number of correctly classified data. Note that the classification accuracy used in supervised setting is consistent with cluster accuracy and is widely used for performance comparison in unsupervised case as in~\cite{SCAN:van2020scan}. In this work, ACC is used to evaluate the model's ability to provide semantic meaningful clusters. 

\begin{table*}[t]
    % \vspace{-0.3cm}
    \centering
  \scalebox{.9}{
    \begin{tabular}{ccccccccc|cccc}
        \hline
        Datasets & \multicolumn{8}{c}{\textbf{CIFAR-100}} & \multicolumn{4}{|c}{\textbf{ImageNet}} \\
        \hline
        Step size& \multicolumn{2}{c}{5} & \multicolumn{2}{c}{10} & \multicolumn{2}{c}{20} & \multicolumn{2}{c}{50} & \multicolumn{2}{|c}{10} & \multicolumn{2}{c}{100}\\
        \hline
        ACC & Avg & Last & Avg & Last & Avg & Last & Avg & Last & Avg & Last & Avg & Last\\
        \hline
        \hline
        Scratch  & 0.106 & 0.038& 0.095& 0.015& 0.122 & 0.038 & 0.226 & 0.226 & 0.282 & 0.158 & 0.069 & 0.023 \\
        PCA & 0.156 & 0.085& 0.143& 0.061& 0.171 & 0.083 & 0.287 & 0.287 & 0.308 & 0.175 & / & / \\
        FFE & 0.459 & 0.338& 0.399& 0.281& 0.401 & 0.323 & 0.392 & 0.392 & 0.757 & 0.620 & 0.405 & 0.275 \\
        UPL-10 & 0.498 & 0.376 & 0.415 & 0.293 & 0.430 &0.320 & 0.401 & 0.401  & 0.797 & 0.653 & 0.446 & 0.294 \\
        UPL-20 & 0.523 & 0.394 & 0.422 & 0.296 & 0.445 &0.339 & 0.413 & 0.413  & 0.816 & 0.699 & 0.458 & 0.311 \\
        UPL-30 & 0.513 & 0.383 & 0.435 & 0.324 & 0.459 &0.364 & 0.433 & 0.433  & 0.832 & 0.705 & 0.460 & 0.332 \\
        Ours &  \textbf{0.558} & \textbf{0.426} & \textbf{0.482} & \textbf{0.368} & \textbf{0.486} & \textbf{0.397} & \textbf{0.495} & \textbf{0.495} & \textbf{0.849} & \textbf{0.722} & \textbf{0.471} & \textbf{0.342}  \\
        \hline
    \end{tabular}
    }
    \vspace{-0.2cm}
    \caption{\textbf{Ablation study for different approaches to obtain pseudo labels on CIFAR-100 and ImageNet} in terms of average ACC (Avg) and last step ACC (Last). The best results are marked in bold. }
    \label{tab:ablation:clustering}
\end{table*}

\subsection{Implementation Detail}
Our implementation is based on Pytorch~\cite{pytorch} and we use ResNet-32 for CIFAR-100 and ResNet-18 for ImageNet. The ResNet implementation follows the setting as suggested in~\cite{RESNET}. The setting of incorporated existing approaches follows their own repositories. We select $q = 20$ exemplars per cluster to construct exemplar set and arrange classes using identical random seed (1993) with benchmark supervised experiment protocol~\cite{ICARL}. We ran five times for each experiment and the average performance is reported. 

\subsection{Incorporating with Supervised Approaches} 
% \subsection{Results of Incorporating into Supervised Approaches} 
\label{results: compare with supervised}
In this part, our method is evaluated when incorporated into existing supervised approaches including \textbf{LWF}~\cite{LWF}, \textbf{ICARL}~\cite{ICARL}, \textbf{EEIL}~\cite{EEIL}, \textbf{LUCIR}~\cite{rebalancing}, \textbf{WA}~\cite{mainatining} and \textbf{ILIO}~\cite{ILIO}, which are representative methods from all \textit{Regularization based}, \textit{Bias-correction based} and \textit{Rehearsal based} categories as described in Section~\ref{related work}. Note that \textbf{ILIO} is implemented in online scenario where each data is used only once to update model while others are implemented in offline. We embed the pseudo label as illustrated in Section~\ref{method_our_overview} to evaluate the performance of selected approaches in unsupervised mode. \textit{E.g.} \textbf{ICARL + Ours} denotes the implementation of \textbf{ICARL} in unsupervised mode by incorporating with our proposed method. Table~\ref{tab:incorpor} summarizes results in terms of last step ACC (Last) and average ACC (Avg) calculated by averaging ACC for all incremental steps, which shows overall performance for the entire continual learning procedure. We also report the performance difference $\Delta = w/ - w/o$ and observe only small degradation by comparing unsupervised results with supervised results. In addition, we calculate the average accuracy drop by $Avg(\Delta) = Avg(w/) - Avg(w/o)$ for each incremental step corresponds to each method. The $Avg(\Delta)$ ranges from $[0.015, 0.295]$ with an average of $0.114$. Our method can work well with but not limited to these selected representative methods and we achieve competitive performance in unsupervised scenario without requiring human annotated labels during continual learning phase. Figure~\ref{fig:result on cifar} shows cluster accuracy for each incremental step on CIFAR-100. (More results and discussion are available in the \textit{Appendix})

% \vspace{-0.2cm}
\subsection{Ablation Study} 
% \vspace{-0.2cm}
\label{exp:ablation}
We conduct extensive experiments to \textbf{1)} analyze the unsupervised exemplar selection step as described in Section~\ref{method:continual learning} by varying the number of exemplars per class and compare the results with random selection. \textbf{2)} Study the impacts of different methods that can be used to extract feature for clustering to obtain pseudo label during continual learning. For both experiments, we first construct our baseline method denoted as \textbf{Ours}, which uses distillation loss as described in Equation~\ref{eq:cdloss} and exemplars from learned tasks as described in Algorithm~\ref{alg:exp_selection}. (see implementation detail in \textit{Appendix})

\begin{figure}[t]
\begin{center}
%   \vspace{-0.5cm}
  \includegraphics[width=.9\linewidth]{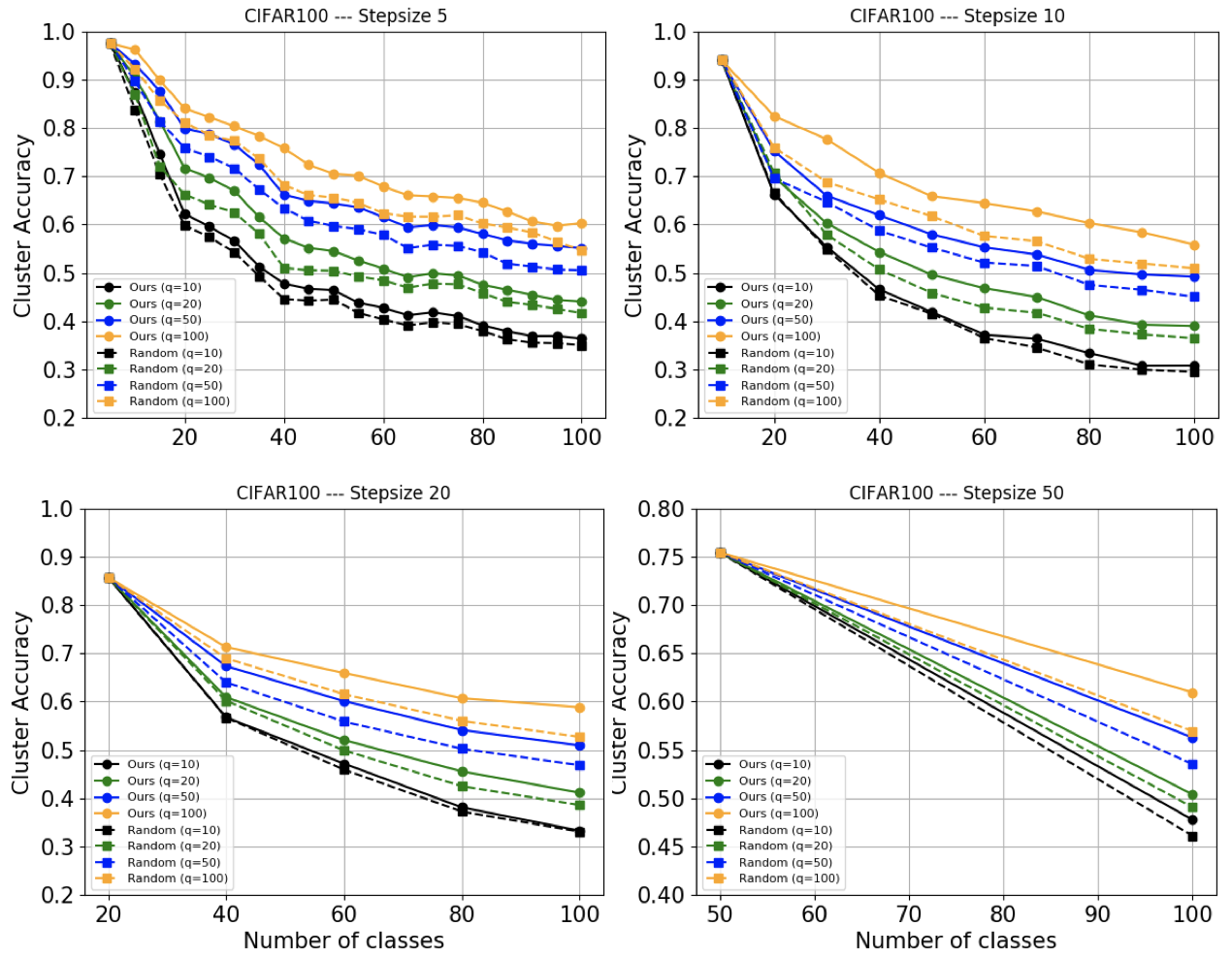}
  \vspace{-0.3cm}
  \caption{Results on CIFAR-100 by varying target exemplar size $q \in \{10, 20, 50, 100\}$ and comparison with random selection.}
%   \vspace{-.5cm}
  \label{fig:exp}
\end{center}
\end{figure}

For part \textbf{1)}, we vary the target number of exemplars per class $q \in \{10, 20, 50, 100\}$ and compare the results with random exemplar selection from each generated cluster, denoted as \textbf{Random}. The results on CIFAR-100 are shown in Figure~\ref{fig:exp}. We observe that the overall performance will be improved by increasing $q$ even using randomly selected exemplars. In addition, our proposed method, which selects exemplars based on cluster mean, outperforms \textbf{Random} by a larger margin when $q$ becomes larger. 

For part \textbf{2)}, we compare our method using the updated model from last incremental step as feature extractor as illustrated in Algorithm~\ref{alg:unsupervised continual learning} with i) \textbf{Scratch}: apply a scratch model with the same network architecture as feature extractor, ii) \textbf{PCA}: directly apply PCA algorithm~\cite{wold1987principal_PCA} on input images to obtain feature embeddings for clustering, iii) \textbf{Fixed Feature Extractor (FFE)}: use model $\textbf{h}_1$ as described in Section~\ref{method:obtain pseudo label} as the fixed feature extractor for the entire continual learning process, iv) \textbf{Updated Pseudo Label (UPL-K)}: iteratively update model and perform clustering within each incremental step as proposed in~\cite{deep_clustering}, where $K$ indicates how frequently we update the pseudo label \textit{e.g.} UPL - 10 means we update pseudo label for every 10 epochs. All these variants are modified based on our baseline method. Results are summarized in Table~\ref{tab:ablation:clustering}. The scratch method provides lower bound performance and FFE outperforms PCA by a large margin, showing the advanced ability of using deep models to extract more discriminative feature for clustering. Note that we did not perform PCA on ImageNet-1000 as it takes quite a long time for computation. Comparing UPL-K with $K = 0, 10, 20, 30$ ($K=0$ is Ours), we observe that if the updating frequency increases ($K$ decreases), the overall performance degrades. As discussed in Section~\ref{method:obtain pseudo label}, different from unsupervised representation learning that uses a model from scratch, in continual learning we also need to preserve the learned knowledge for all classes seen so far and update pseudo label repeatedly will accelerate the catastrophic forgetting, resulting in the performance drop.
% Figure~\ref{fig:upl} visualizes the test ACC of UPL-K on CIFAR-100 for each incremental step using step size 10 and 20. 

% \begin{figure}[t]
% \begin{center}
% %   \vspace{-0.5cm}
%   \includegraphics[width=1.\linewidth]{fig_experiment/FIG_UPL.png}
%   %\vspace{-0.2cm}
%   \caption{ACC results of \textbf{UPL-K} on CIFAR-100 with K = 0 (Ours), 10, 20, and 30. (Best viewed in color)}
% %   \vspace{-.5cm}
%   \label{fig:upl}
% \end{center}
% \end{figure}

\vspace{-0.2cm}
\section{Conclusion}
\vspace{-0.2cm}
\label{conclusion}
In summary, we explore a novel problem of unsupervised continual learning under class-incremental setting where the objective is to learn new classes incrementally while providing semantic meaningful clusters on all classes seen so far. We proposed a simple yet effective method using pseudo labels obtained based on cluster assignments to learn from unlabeled data for each incremental step. We introduced a new experimental protocol and evaluate our method on benchmark image classification datasets including CIFAR-100 and ImageNet (ILSVRC). We demonstrate that our method can be easily embedded with various existing supervised approaches implemented under both online and offline modes to achieve competitive performance in unsupervised scenario. Finally, we show that our proposed exemplar selection method works effectively without requiring ground truth and iteratively updating pseudo labels will cause performance degradation under continual learning context.

\bibliographystyle{named}
\bibliography{ijcai21}

\appendix

\section{Implementation Detail For Our Baseline Method}
\label{sec:baseline}
\begin{figure}[t]
\begin{center}
%   \vspace{-0.5cm}
  \includegraphics[width=1.\linewidth]{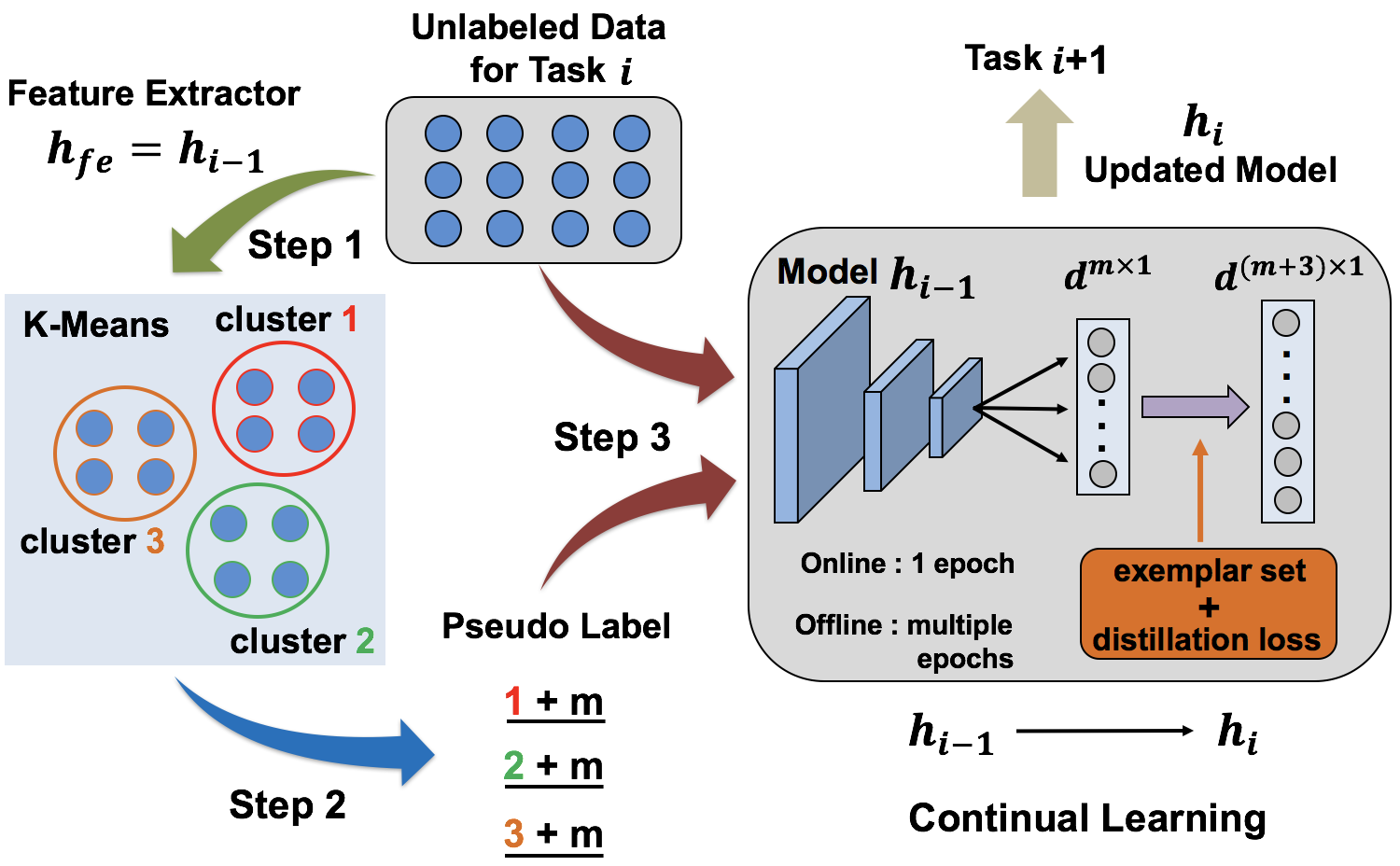}
  %\vspace{-0.2cm}
  \caption{\textbf{Overview of our baseline solution to learn the new task $i$.} \textbf{h} refers to the model in different steps and $\textbf{m}$ denotes the number of learned classes so far after task $i-1$. Firstly, we apply $\textbf{h}_{i-1}$ (except the last fully connected layer) to extract feature embeddings used for K-means clustering where the number $\textbf{1}$, $\textbf{2}$, $\textbf{3}$ denote the corresponding cluster assignments. In step 2 we obtain the pseudo label $\textbf{1+m}$, $\textbf{2+m}$, $\textbf{3+m}$ respectively. Finally in step 3, the unlabeled data with pseudo label is used together for continual learning. }
  \vspace{-.5cm}
  \label{fig:method}
\end{center}
\end{figure}

In the paper, we conduct extensive experiments in \textbf{Section 5.5}\footnote{Section, Table and Figure references marked in bold can be found in the submitted paper} using a baseline solution denoted as \textbf{Ours}. The overview of our baseline method is shown in Figure~\ref{fig:method}, which incorporates pseudo labels introduced in \textbf{Section 4.1} with knowledge distillation loss as in \textbf{Equation 3} and exemplar replay as described in \textbf{Algorithm 2}. Thus, the difference between our baseline solution and \textbf{EEIL} is that we exclude data augmentation and balanced fine-tuning steps. 

We apply ResNet-32 for CIFAR and ResNet-18 for ImageNet, which keeps same with in \textbf{Section 5.3}. We use batch size of 128 with initial learning rate of 0.1. SGD optimizer is applied with weight decay of 0.00001. We train 120 epochs for each incremental step and the learning rate is decreased by $1/10$ for every 30 epochs. We perform each experiment 5 times and the average results are reported in \textbf{Table 2} and \textbf{Figure 3}.

\section{Additional Experimental Results}
\label{sec:additioanla exp results}
In this section, we show additional experiment results for (1) cluster accuracy for each incremental step on ImageNet. (2) Analysis of performance drop compared with supervised results corresponds to \textbf{Table 1}. (3) Impact of different clustering algorithms by comparing K-means clustering with Gaussian Mixture Models (GMM). (4) Results in terms of other evaluation metrics including \textit{Normalized Mutual Information} (NMI) and \textit{Adjusted Rand Index} (ARI). Both experiments are implemented by using our \textbf{baseline} solution as described in Section~\ref{sec:baseline} on test data of CIFAR-100 with step size 5, 10, 20, 50.

\begin{figure}[t]
\begin{center}
%   \vspace{-0.5cm}
  \includegraphics[width=1.\linewidth]{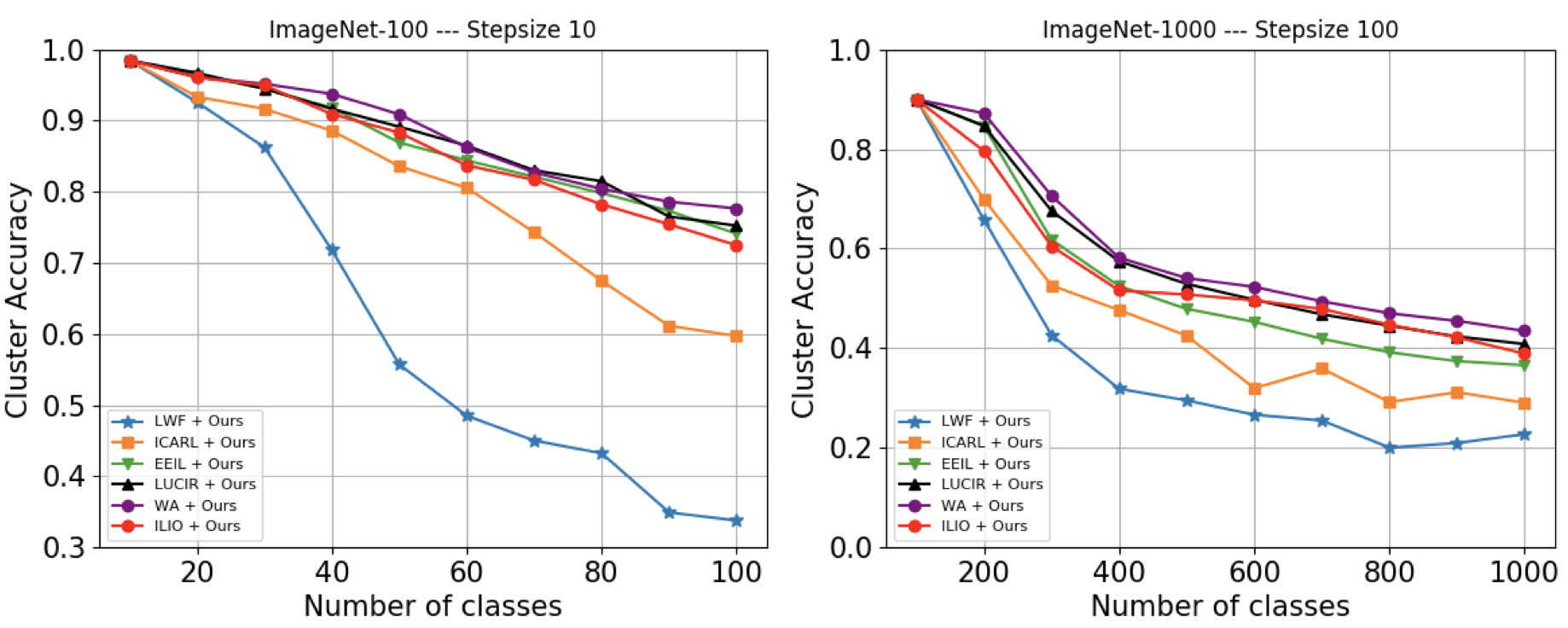}
  \vspace{-0.6cm}
  \caption{\textbf{Results on ImageNet} with step size (a) 10 on ImageNet-100 and (b) 100 on ImageNet-1000.(Best viewed in color)}
%   \vspace{-.3cm}
  \label{fig:imagenet}
\end{center}
\end{figure}

\begin{table}[t]
    % \vspace{-0.3cm}
    \centering
    \scalebox{.9}{
    \begin{tabular}{|ccccccc|}
        \hline
        Datasets & \multicolumn{4}{c}{\textbf{CIFAR-100}} & \multicolumn{2}{c|}{\textbf{ImageNet}} \\
        \hline
        Step size& 5 & 10 & 20 & 50 & 10 & 100\\
        % \hline
        % ACC & \multicolumn{1}{c}{Avg} & \multicolumn{1}{c}{Last} & \multicolumn{1}{c}{Avg} & \multicolumn{1}{c}{Last} & \multicolumn{1}{c}{Avg} & \multicolumn{1}{c}{Last} & \multicolumn{1}{c}{Avg} &
        % \multicolumn{1}{c}{Last} & \multicolumn{1}{|c}{Avg} &
        % \multicolumn{1}{c}{Last} & \multicolumn{1}{c}{Avg} &
        % \multicolumn{1}{c}{Last} \\
        \hline
        \hline
        LWF & -0.071 & -0.091& -0.086& -0.095& -0.033 & -0.211\\
         ICARL & -0.084 & -0.132& -0.158& -0.108& -0.043 & -0.197 \\
        EEIL & -0.071 & -0.131& -0.131& -0.088& -0.040 & -0.199 \\
        LUCIR & -0.015 & -0.104& -0.131& -0.111& -0.037 & -0.293 \\
        WA & -0.034 & -0.110 & -0.121 & -0.092 & -0.037 & -0.295 \\
        ILIO & -0.123 & -0.140& -0.134& -0.106& -0.057 & -0.178 \\
        
        \hline
    \end{tabular}
    }
    \caption{\textbf{Summary of performance degradation $Avg(\Delta) = Avg(w/) - Avg(w/o)$ in terms of average accuracy \textit{Avg}}. }
    \label{tab:drop}
\end{table}

\begin{figure*}[t!]
\centering
\begin{minipage}[t]{0.22\linewidth}
    \centering
    \includegraphics[width=4.cm,height=3.5cm]{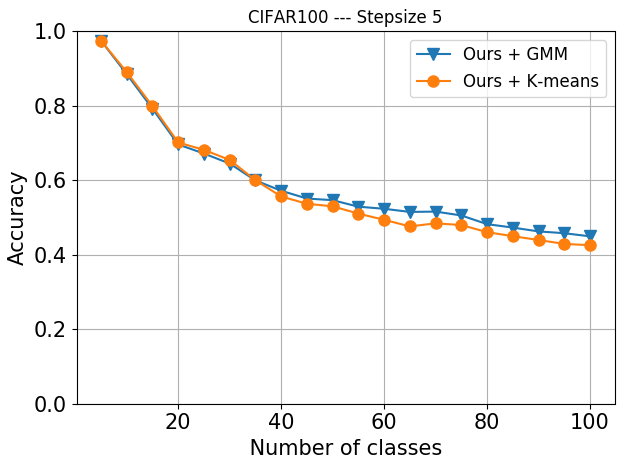}
    \parbox{15.5cm}{\small \hspace{2.cm}(a)}
\end{minipage}
\hspace{2ex}
\begin{minipage}[t]{0.22\linewidth}
    \centering
    \includegraphics[width=4.cm,height=3.5cm]{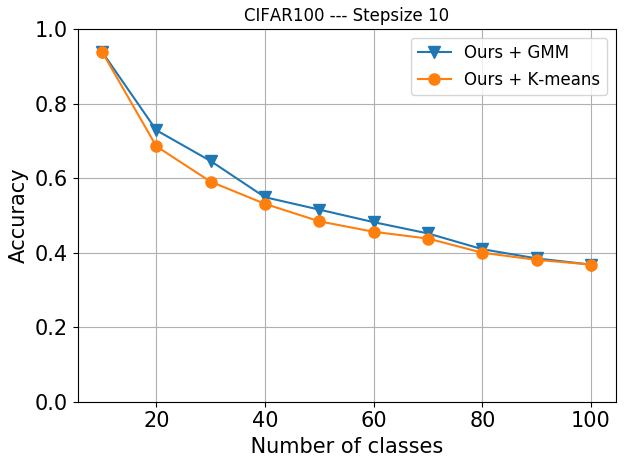}
    \parbox{15.5cm}{\small \hspace{2.cm}(b)}
\end{minipage}
\hspace{2ex}
\begin{minipage}[t]{0.22\linewidth}
    \centering
    \includegraphics[width=4.cm,height=3.5cm]{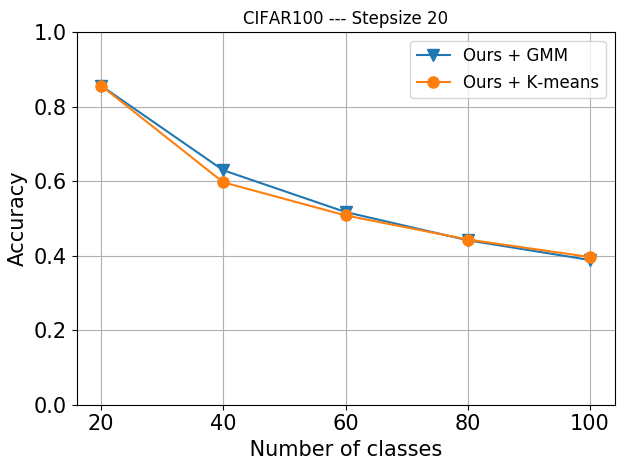}
    \parbox{15.5cm}{\small \hspace{2.cm}(c)}
\end{minipage}
\hspace{2ex}
\begin{minipage}[t]{0.22\linewidth}
    \centering
    \includegraphics[width=4.cm,height=3.5cm]{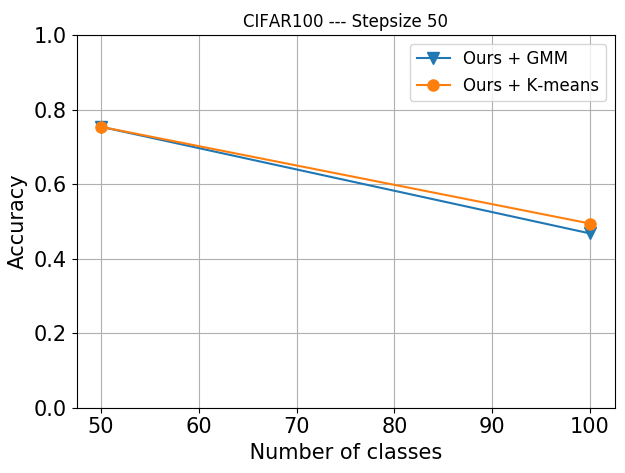}
    \parbox{15.5cm}{\small \hspace{2.cm}(d)}
\end{minipage}
\begin{center}
\vspace{-0.3cm}
\caption{Results of test data on CIFAR-100 for comparing K-means and GMM with incremental step size (a) 5, (b) 10, (c) 20 and (d) 50. The Accuracy refers to cluster accuracy (ACC). (Best viewed in color)  }
\label{fig:part1}
\end{center}
\end{figure*}

\begin{figure*}[t!]
\vspace{-0.5cm}
\centering
\begin{minipage}[t]{0.22\linewidth}
    \centering
    \includegraphics[width=4.cm,height=3.5cm]{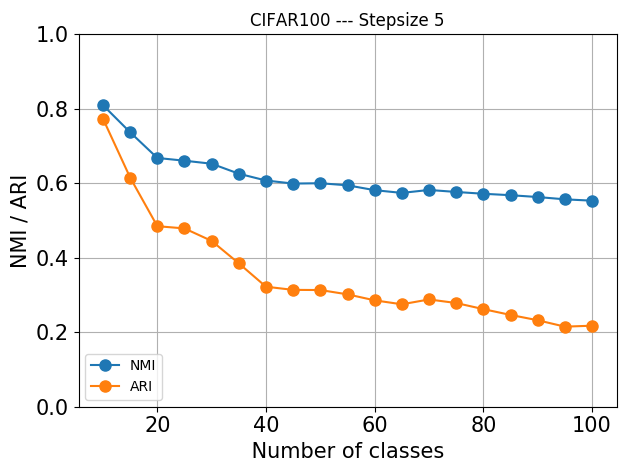}
    \parbox{15.5cm}{\small \hspace{2.cm}(a)}
\end{minipage}
\hspace{2ex}
\begin{minipage}[t]{0.22\linewidth}
    \centering
    \includegraphics[width=4.cm,height=3.5cm]{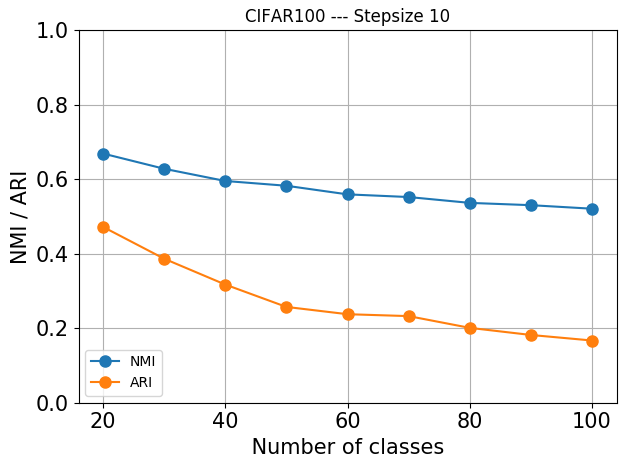}
    \parbox{15.5cm}{\small \hspace{2.cm}(b)}
\end{minipage}
\hspace{2ex}
\begin{minipage}[t]{0.22\linewidth}
    \centering
    \includegraphics[width=4.cm,height=3.5cm]{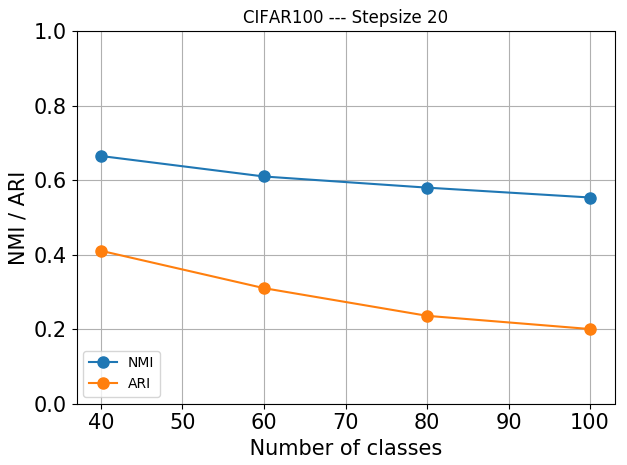}
    \parbox{15.5cm}{\small \hspace{2.cm}(c)}
\end{minipage}
\hspace{2ex}
\begin{minipage}[t]{0.22\linewidth}
    \centering
    \includegraphics[width=4.cm,height=3.5cm]{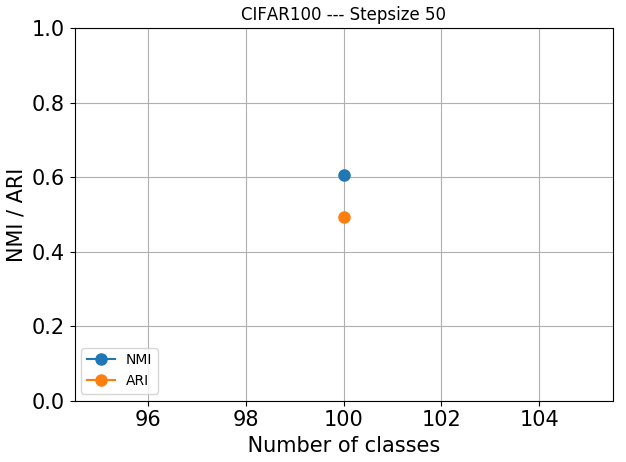}
    \parbox{15.5cm}{\small \hspace{2.cm}(d)}
\end{minipage}
\begin{center}
\caption{Offline results in terms of NMI and ARI of test data on CIFAR-100 with incremental step size (a) 5, (b) 10, (c) 20 and (d) 50. The Accuracy refers to cluster accuracy (ACC). Note that only unsupervised incremental results (starting from task 2) are reported in this part. (Best viewed in color)  }
\label{fig:part2}
\end{center}
\end{figure*}

\subsection{Results on ImageNet}
\label{sec: imagenet results}
The cluster accuracy evaluated after each incremental step on CIFAR-100 with different step sizes are
shown in \textbf{Figure 2} and in this part we provide the results on ImageNet with step size $10$ and $100$ as shown in Figure~\ref{fig:imagenet}.

\subsection{Analysis of Performance Drop}
\label{sec:performance drop}
In \textbf{Section 5.4}, we incorporate our method with existing supervised approaches and results are shown in \textbf{Table 1}. In this part, we further investigate the performance degradation in unsupervised scenario. Specifically, we calculate the average accuracy
drop by $Avg(\Delta) = Avg(w/) - Avg(w/o)$ for each incremental step corresponds to each method. The results are shown in Table~\ref{tab:drop} where $Avg(\Delta)$ ranges from $[0.015, 0.295]$ with an average of $0.114$. We notice that the performance degradation for each incremental step do not vary a lot for different approaches. Therefore, the methods with higher accuracy in supervised case are more likely to achieve higher performance in unsupervised scenario by incorporating with our pseudo labels. In addition, the performance degradation will increase in online scenario (ILIO) as well as for very large incremental step size (100), which are both challenging cases even in supervised continual learning with human annotations.

\subsection{K-means VS. GMM}
\label{sec:part 1}
As illustrated in \textbf{Section 4}, our proposed method use K-means clustering for illustration purpose to obtain the pseudo labels and to sample exemplars. In this part, we show the results in terms of cluster accuracy (ACC) for each incremental step by comparing K-means with GMMs, which estimates the parameters of each Gaussian distribution through expectation-maximization algorithm. Figure~\ref{fig:part1} shows the ACC results on CIFAR-100 with step size 5, 10, 20 and 50. We observe only small performance difference for all incremental steps, which shows that the choice of clustering methods are not crucial for our proposed method. 

\subsection{Results Evaluated By NMI and ARI}
\label{sec:part2}
In the paper we use cluster accuracy (ACC) to evaluate the model's ability to provide semantic meaningful cluster on test data as illustrated in \textbf{Section 5.2}. In this part, in addition to ACC, we also provide results in terms of NMI and ARI to measure the quality of obtained semantic meaningful clusters. Let $A$ and $B$ refer to ground truth labels and generated cluster assignments. 

NMI measures the shared information between two clustering assignments A and B by $$NMI(A,B) = \frac{I(A,B)}{\sqrt{H(A)H(B)}}$$ where $H(\sbullet[.5])$ and $I(\sbullet[.5])$ denote entropy and mutual information, respectively. 

ARI is defined by $$ARI(A,B) = \frac{\sum_{i,j}      \begin{pmatrix}
    N_{i,j} \\
    2
  \end{pmatrix} -\frac{\sum_{i} \begin{pmatrix}
    N_{i} \\
    2
  \end{pmatrix}\sum_{j}      \begin{pmatrix}
    N_{j} \\
    2
  \end{pmatrix}}{\begin{pmatrix}
    N \\
    2
  \end{pmatrix}}}{\frac{1}{2}[\sum_{i} \begin{pmatrix}
    N_{i} \\
    2
  \end{pmatrix} + \sum_{j} \begin{pmatrix}
    N_{j} \\
    2
  \end{pmatrix}] - \frac{\sum_{i} \begin{pmatrix}
    N_{i} \\
    2
  \end{pmatrix}\sum_{j} \begin{pmatrix}
    N_{j} \\
    2
  \end{pmatrix}}{\begin{pmatrix}
    N \\
    2
  \end{pmatrix}}}$$ where $\begin{pmatrix}
    N \\
    2
  \end{pmatrix}$ refers to binomial coefficients and $N$ is the total number of data in the cluster. $N_i$ and $N_j$ denote number of data with cluster assignment $C_i$ in $B$ and the number of data with class label $C_j^{\star}$ in $A$, respectively. $N_{i,j}$ is the number of data with the class label $C_j^{\star} \in A$ assigned to cluster assignment $C_i \in B$.

NMI and ARI ranges from $[0, 1]$ and 1 means a perfect match. Figure~\ref{fig:part2} shows the results on CIFAR-100 with incremental step size 5, 10, 20, and 50 using our baseline method. We observe consistent performance compared with using cluster accuracy (ACC) as metric as shown in \textbf{Figure 2} in paper. Note that both results are reported starting from the second task (first incremental step).

\end{document}